\pdfoutput=1

\documentclass{article}
\usepackage{ijcai22}

\usepackage{times}
\usepackage{soul}
\usepackage{url}
\usepackage[hidelinks]{hyperref}
\usepackage[utf8]{inputenc}
\usepackage[small]{caption}
\usepackage{graphicx}
\usepackage{amsmath}
\usepackage{amsthm}
\usepackage{amssymb}
\usepackage{booktabs}
\usepackage{algorithm}
\usepackage{algorithmic}
\usepackage{xcolor}
\usepackage{mathtools}
\usepackage{verbatim}
\usepackage{subfigure}
\usepackage{diagbox}
\usepackage{csquotes}
\usepackage{multirow}
\usepackage{cprotect}

\newcommand{\figref}[1]{Fig.~\ref{#1}}
\newcommand{\tabref}[1]{Tab.~\ref{#1}}
\newcommand{\secref}[1]{Sec.~\ref{#1}}

\usepackage{tabularx}
\newcolumntype{P}[1]{>{\centering \arraybackslash}p{#1}}
\newcolumntype{L}{X}
\newcolumntype{C}{>{\centering \arraybackslash}X}
\newcolumntype{R}{>{\raggedright \arraybackslash}X}

\title{SparseTT: Visual Tracking with Sparse Transformers}

\author{
Zhihong Fu\and
Zehua Fu\and
Qingjie Liu\footnote{Contact Author}\and
Wenrui Cai\And
Yunhong Wang
\affiliations
State Key Laboratory of Virtual Reality Technology and Systems, Beihang University, Beijing, China\\
Hangzhou Innovation Institute, Beihang University\\
\emails
\{fuzhihong, zehua\_fu, qingjie.liu, wenrui\_cai, yhwang\}@buaa.edu.cn
}

\begin{document}

\maketitle

\begin{abstract}
\vspace{-0.7em}
Transformers have been successfully applied to the visual tracking task and significantly promote tracking performance.
The self-attention mechanism designed to model long-range dependencies is the key to the success of Transformers.
However, self-attention lacks focusing on the most relevant information in the search regions, making it easy to be distracted by background.
In this paper, we relieve this issue with a sparse attention mechanism by focusing the most relevant information in the search regions, which enables a much accurate tracking.
Furthermore, we introduce a double-head predictor to boost the accuracy of foreground-background classification and regression of target bounding boxes, which further improve the tracking performance.
Extensive experiments show that, without bells and whistles, our method significantly outperforms the state-of-the-art approaches on LaSOT, GOT-10k, TrackingNet, and UAV123, while running at 40 FPS.
Notably, the training time of our method is reduced by 75\% compared to that of TransT.
The source code and models are available at \url{https://github.com/fzh0917/SparseTT}.
\vspace{-1.0em}
\end{abstract}

\section{Introduction}
\begin{figure}[t]
    \centering
    \includegraphics[width=0.49\textwidth]{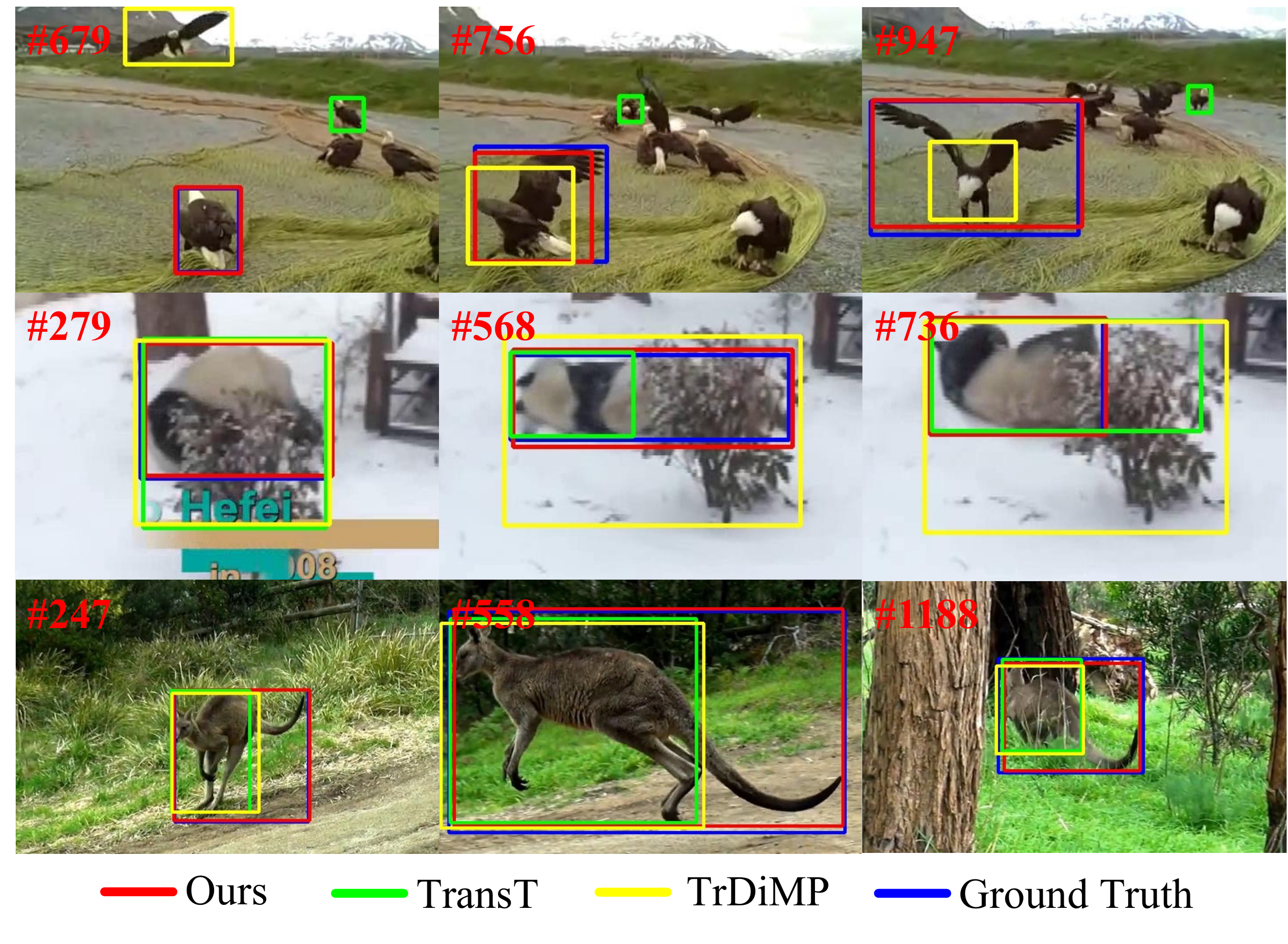}
    \vspace{-2.0em}
    \caption{Visualized comparisons of our method with excellent trackers TransT~\protect\cite{chen2021transformer} and TrDiMP~\protect\cite{wang2021transformer}. Our method enables the bounding boxes of targets to be more accurate even under severe target deformation, partial occlusion, and scale variation. Zoom in for better view.}
    \label{fig:illustration_at_the_first_page}
 \vspace{-1.3em}
\end{figure}

\begin{figure*}[t]
    \centering
    \includegraphics[width=0.98\textwidth]{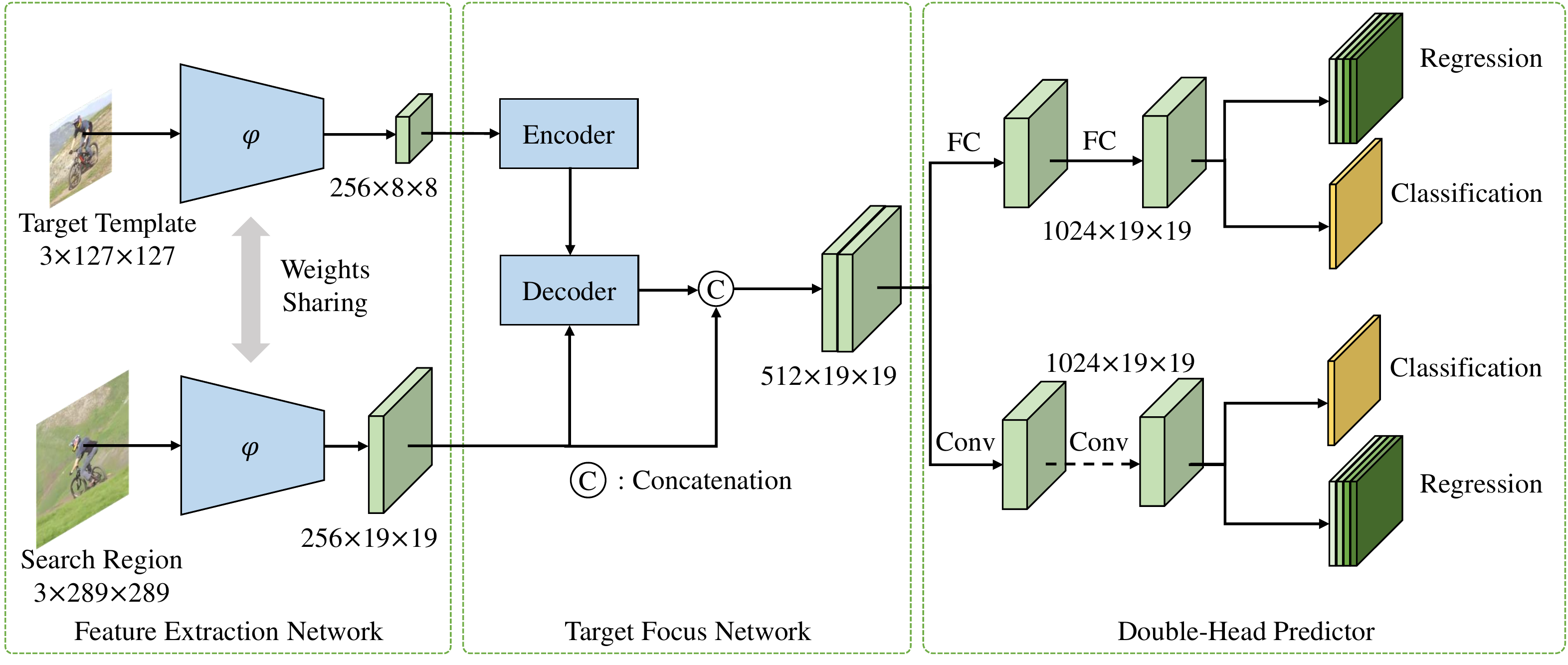}
    \vspace{-1.0em}
    \caption{The architecture of our method.}
    \label{fig:architecture}
 \vspace{-1.0em}
\end{figure*}

Visual tracking aims to predict the future states of a target given its initial state.
It is applicable broadly, such as human-computer interactions, video surveillance, and autonomous driving.
Most of the existing methods address the tracking problem with sequence prediction frameworks where they estimate the current state based on the initial and the previous states.
Thus, it is important to give accurate states in every time slice otherwise errors accumulate and will lead to tracking failure.
Significant efforts have been devoted to improving the tracking accuracy, i.e., the accuracy of the target bounding boxes.
However, challenges such as target deformation, partial occlusion, and scale variation are still huge obstacles ahead hindering them from perfect tracking.
The reason may be that most of these methods adopt cross-correlation operation to measure similarities between the target template and the search region, which may trap into local optimums.
Recently, TransT~\cite{chen2021transformer} and DTT~\cite{yu2021high} improve the tracking performance by replacing the correlation with Transformer~\cite{Vaswani2017AttentionIA}.
However, building trackers with Transformers will lead to a new problem: the global perspective of self-attention in Transformers causes the primary information (such as targets in search regions) under-focused, but the secondary information (such as background in search regions) over-focused, making the edge region between the foreground and background blurred, and thus degrade the tracking performance.

In this paper, we attack this issue by concentrating on the most relevant information of the search region, which is realized with a  sparse Transformer.
Different from vanilla Transformers used in previous works~\cite{chen2021transformer,yu2021high}, sparse Transformer is designed to focus on primary information, enabling the targets to be more discriminative and the bounding boxes of targets to be more accurate even under severe target deformation, partial occlusion, scale variation, and so on, as shown in \figref{fig:illustration_at_the_first_page}.

Summarily, the main contributions of this work are three-fold.
\begin{itemize}
\item{We present a target focus network that is capable of focusing on the target of interest in the search region and highlighting the features of the most relevant information for better estimating the states of the target.}
\item{We propose a sparse Transformer based siamese tracking framework that has a strong ability to deal with target deformation, partial occlusion, scale variation, and so on.}
\item{Extensive experiments show that our method outperforms the state-of-the-art approaches on LaSOT, GOT-10k, TrackingNet, and UAV123, while running at 40 FPS, demonstrating the superiority of our method.}
\end{itemize}

\section{Related Work}
\paragraph{Siamese Trackers.}
In siamese visual trackers, cross-correlation, commonly used to measure the similarity between the target template and the search region, has been extensively studied for visual tracking.
Such as naive cross-correlation~\cite{bertinetto2016fully}, depth-wise cross-correlation~\cite{li2019siamrpn++,xu2020siamfc++}, pixel-wise cross-correlation~\cite{yan2021alpha}, pixel to global matching cross-correlation~\cite{liaopg}, etc.
However, cross-correlation performs local linear matching processes, which may fall into local optimum easily~\cite{chen2021transformer}.
And furthermore, the cross-correlation captures relationships and thus corrupts semantic information of the inputted features, which is adverse to accurate perception of target boundaries.
Most siamese trackers still have difficulties dealing with target deformation, partial occlusion, scale variation, etc.

\paragraph{Transformer in Visual Tracking.}
Recently, Transformers have been successfully applied to visual tracking field.
Borrowing inspiration from DETR~\cite{carion2020end}, STARK~\cite{Yan_2021_ICCV} casts target tracking as a bounding box prediction problem and solve it with an encoder-decoder transformer, in which the encoder models the global spatio-temporal feature dependencies between targets and search regions, and the decoder learns a query embedding to predict the spatial positions of the targets.
It achieves excellent performance on visual tracking.
TrDiMP~\cite{wang2021transformer} designs a siamese-like tracking pipeline where the two branches are built with CNN backbones followed by a Transformer encoder and a Transformer decoder, respectively.
The Transformers here are used to enhance the target templates and the search regions.
Similar to previous siamese trackers, TrDiMP applies cross-correlation to measure similarities between the target templates and the search region, which may impede the tracker from high-performance tracking.
Noticing this shortcoming, TransT~\cite{chen2021transformer} and DTT~\cite{yu2021high} propose to replace cross-correlation with Transformer, thereby generating fused features instead of response scoress.
Since fused features contain rich semantic information than response scores, these methods reach much accurate tracking than previous siamese trackers.

Self-attention in Transformers specializes in modeling long-rang dependencies, making it good at capturing global information, however, suffering from a lack of focusing on the most relevant information in the search regions.
To further boost Transformer trackers, we alleviate the aforementioned drawback of self-attention with a sparse attention mechanism.
The idea is inspired by~\cite{zhao2019explicit}.
We adapt the sparse Transformer in~\cite{zhao2019explicit} to suit the visual tracking task and propose a new end-to-end siamese tracker with an encoder-decoder sparse Transformer.
Driven by the sparse attention mechanism, the sparse Transformer focuses on the most relevant information in the search regions, thus suppressing distractive background that disturbs the tracking more efficiently.

\section{Method}
We propose a siamese architecture for visual tracking, which consists of a feature extraction network, a target focus network, and a double-head predictor, as shown in \figref{fig:architecture}.
The feature extraction network is a weight-shared backbone.
The target focus network built with a sparse Transformer is used to generate target-focused features.
The double-head predictor discriminates foreground from background and outputs bounding boxes of the target.
Note that our method runs at a real-time speed as no online updating in the tracking phase.

\subsection{Target Focus Network}
\begin{figure}[htbp]
    \centering
    \includegraphics[width=0.49\textwidth]{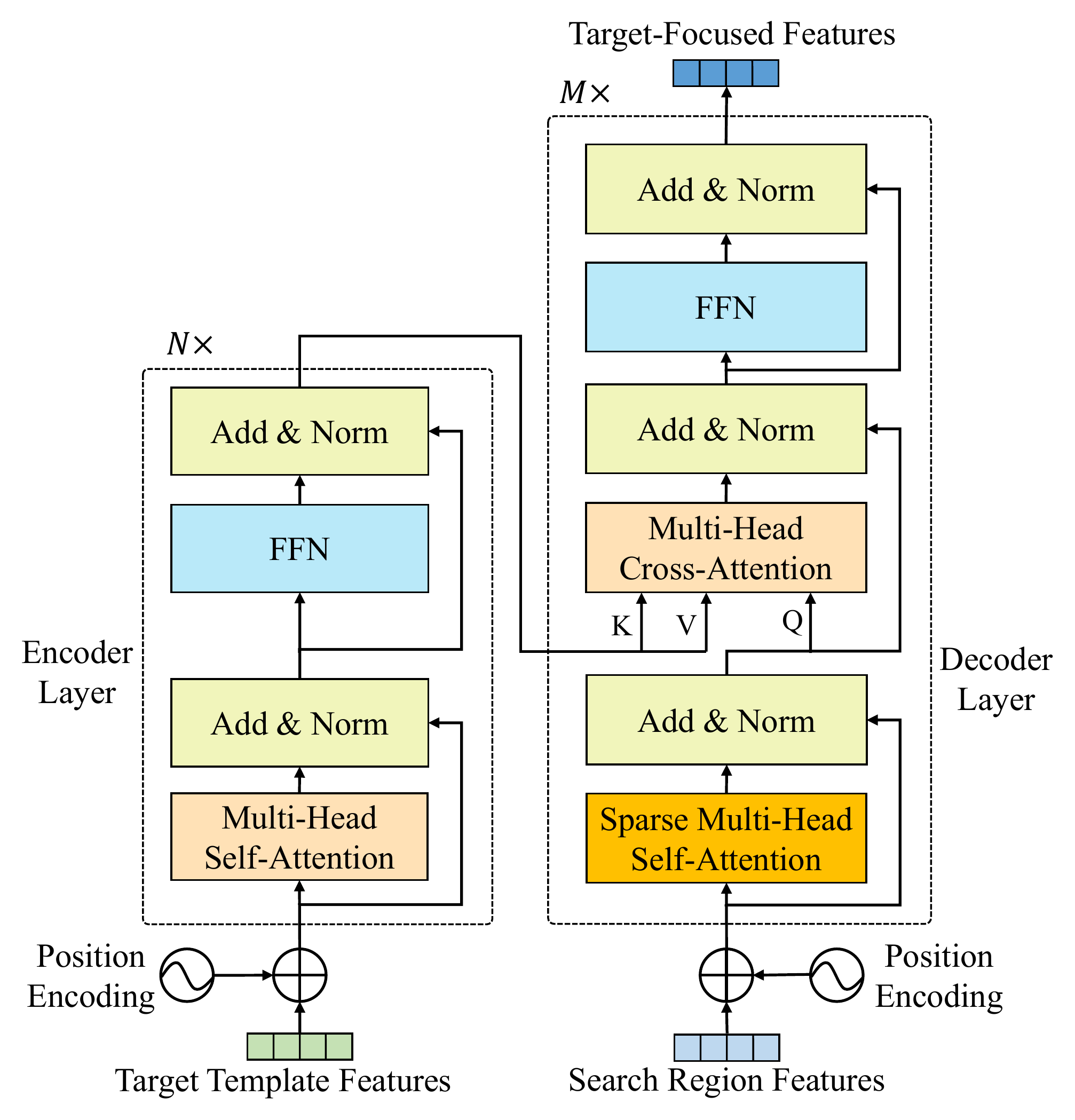}
    \vspace{-2.0em}
    \caption{The architecture of target focus network.}
    \label{fig:sparse_transformer}
 \vspace{-1.0em}
\end{figure}

 The target focus network is built with sparse Transformer, and it has an encoder-decoder architecture, as shown in \figref{fig:sparse_transformer}.
The encoder is responsible for encoding the target template features.
The decoder is responsible for decoding the search region features to generate the target-focused features.

\paragraph{Encoder.} Encoder is an \textbf{important but not essential} component in the proposed target focus network.
It is composed of $N$ encoder layers where each encoder layer takes the outputs of its previous encoder layer as input.
Note that, in order to enable the network to have the perception of spatial position information, we add a spatial position encoding to the target template features, and input the sum to the encoder.
Thus, the first encoder layer takes the target template features with spatial position encoding as input.
In short, it can be formally denoted as:
\begin{equation}\label{eq:encoder}
    \operatorname{encoder}(\boldsymbol{Z}) = \begin{cases}
        f_{enc}^{i}\left(\boldsymbol{Z} + \boldsymbol{P}_{enc}\right), \quad i = 1 \\
        f_{enc}^{i}\left(\boldsymbol{Y}_{enc}^{i-1}\right), \quad 2 \le i \le N \\
\end{cases}
\end{equation}
where $\boldsymbol{Z} \in \mathbb{R}^{H_{t}W_{t} \times C}$ represents the target template features, $\boldsymbol{P}_{enc} \in \mathbb{R}^{H_{t}W_{t} \times C}$ represents the spatial position encoding, $f_{enc}^{i}$ represents the $i$-th encoder layer, $\boldsymbol{Y}_{enc}^{i-1} \in \mathbb{R}^{H_{t}W_{t} \times C}$ represents the output of the $(i-1)$-th encoder layer. $H_{t}$ and $W_{t}$ are the height and width of the feature maps of target templates, respectively.

In each encoder layer, we use multi-head self-attention (MSA) to explicitly model the relations between all pixel pairs of target template features.
Other operations are the same as the encoder layer of vanilla Transformer~\cite{Vaswani2017AttentionIA}.

\paragraph{Decoder.} Decoder is an \textbf{essential} component in the proposed target focus network.
Similar to the encoder, the decoder is composed of $M$ decoder layers.
However, different from the encoder layer, each decoder layer not only inputs the search region features with spatial position encoding or the output of its previous decoder layer, but also inputs the encoded target template features outputted by the encoder.
In short, it can be formally denoted as:
\begin{equation}\label{eq:decoder}
    \operatorname{decoder}(\boldsymbol{X}, \boldsymbol{Y}_{enc}^{N}) = \begin{cases}
        f_{dec}^{i}\left(\boldsymbol{X} + \boldsymbol{P}_{dec}, \boldsymbol{Y}_{enc}^{N}\right), i = 1 \\
        f_{dec}^{i}\left(\boldsymbol{Y}_{dec}^{i-1}, \boldsymbol{Y}_{enc}^{N}\right), 2 \le i \le M \\
    \end{cases}
\end{equation}
where $\boldsymbol{X} \in \mathbb{R}^{H_{s}W_{s} \times C}$ represents the search region features, $\boldsymbol{P}_{dec} \in \mathbb{R}^{H_{s}W_{s} \times C}$ represents the spatial position encoding, $\boldsymbol{Y}_{enc}^{N} \in \mathbb{R}^{H_{t}W_{t} \times C}$ represents the encoded target template features outputted by the encoder, $f_{dec}^{i}$ represents the $i$-th decoder layer, $\boldsymbol{Y}_{dec}^{i-1} \in \mathbb{R}^{H_{s}W_{s} \times C}$ represents the output of $(i-1)$-th decoder layer.
$H_{s}$ and $W_{s}$ are height and width of the feature maps of search regions, respectively.

Different from the decoder layer of vanilla Transformer~\cite{Vaswani2017AttentionIA}, each decoder layer of the proposed sparse Transformer first calculates self-attention on $\boldsymbol{X}$ using sparse multi-head self-attention (SMSA), then calculates cross-attention between $\boldsymbol{Z}$ and $\boldsymbol{X}$ using naive multi-head cross-attention (MCA).
Other operations are the same as the decoder layer of vanilla Transformer~\cite{Vaswani2017AttentionIA}.
Formally, each decoder layer of the proposed sparse Transformer can be denoted as:
\begin{equation}\label{eq:decoder_layer}
\begin{split}
\hat{\boldsymbol{X}} &= \operatorname{Norm}\left(\operatorname{SMSA}\left(\boldsymbol{Y}_{dec}^{i-1}\right) + \boldsymbol{Y}_{dec}^{i-1}\right) \\
\hat{\boldsymbol{Y}}_{dec}^{i} &= \operatorname{Norm}\left(\operatorname{MCA}\left(\hat{\boldsymbol{X}}, \boldsymbol{Y}_{enc}^{N}, \boldsymbol{Y}_{enc}^{N}\right) + \hat{\boldsymbol{X}}\right) \\
\boldsymbol{Y}_{dec}^{i} &= \operatorname{Norm}\left(\operatorname{FFN}\left(\hat{\boldsymbol{Y}}_{dec}^{i}\right) + \hat{\boldsymbol{Y}}_{dec}^{i}\right) \\
\end{split}
\end{equation}

\begin{figure*}[t]
    \centering
    \includegraphics[width=0.99\textwidth]{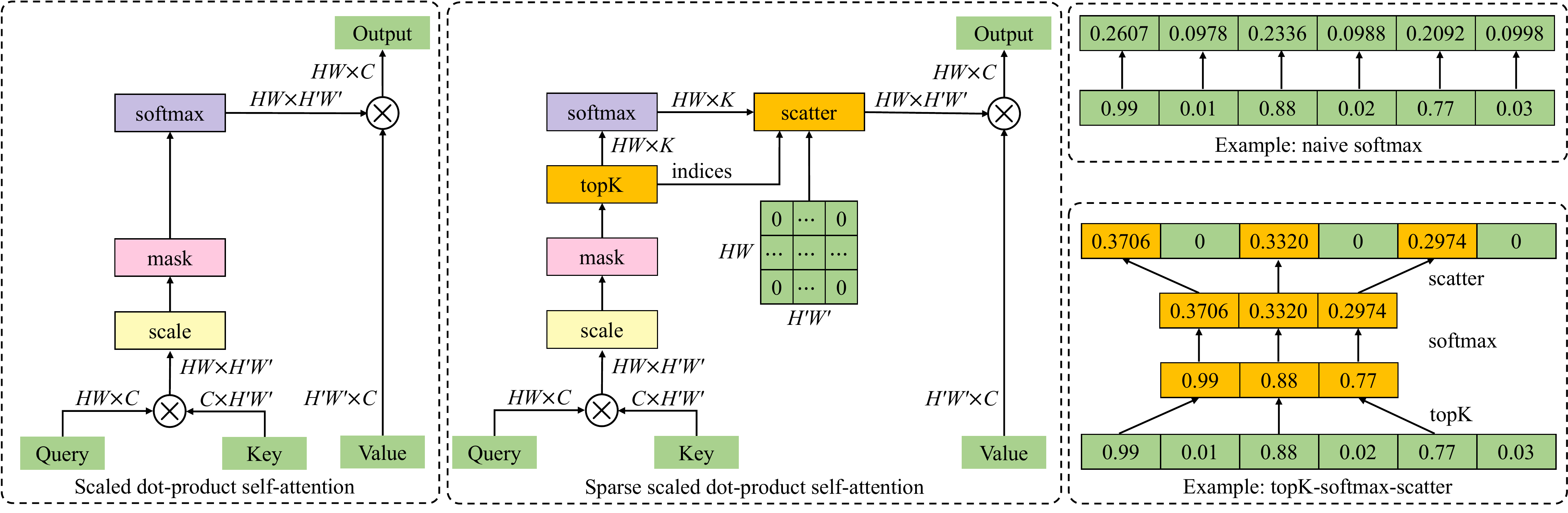}
    \vspace{-1.0em}
    \caption{The left is the illustration of scaled dot-product self-attention in MSA, the middle is the illustration of the sparse scaled dot-product self-attention in SMSA, where the function \texttt{scatter} means filling given values into a 0-value matrix at given indices. The upper right and the lower right are examples of normalizing a row vector of the similarity matrix in naive scaled dot-product attention and sparse scaled dot-product attention, respectively.}
    \label{fig:sparse_dot_product_attention}
 \vspace{-1.0em}
\end{figure*}
\paragraph{Sparse Multi-Head Self-Attention.} Sparse multi-head self-attention is designed to improve the discrimination of foreground-background and to alleviate ambiguity of edge regions of foreground.
Concretely, in the naive MSA, each pixel value of attention features is calculated by all pixel values of the input features, which makes the edge regions of foreground blurred.
In our proposed SMSA, each pixel value of attention features is only determined by $K$ pixel values that are most similar to it, which makes foreground more focused and the edge regions of foreground more discriminative.

Specifically, as shown in the middle of \figref{fig:sparse_dot_product_attention}, given a \textit{\textbf{query}} $\in \mathbb{R}^{HW \times C}$, a \textit{\textbf{key}} $\in \mathbb{R}^{C \times H^{\prime}W^{\prime}}$, and a \textit{\textbf{value}} $\in \mathbb{R}^{H^{\prime}W^{\prime} \times C}$, we first calculate similarities of all pixel pairs between \textit{\textbf{query}} and \textit{\textbf{key}} and mask out unnecessary tokens in the similarity matrix.
Then, different from naive scaled dot-product attention that is shown in the left of \figref{fig:sparse_dot_product_attention}, we only normalize $K$ largest elements from each row of the similarity matrix using \verb|softmax| function.
For other elements, we replace them with $0$.
Finally, we multiply the similarity matrix and \textit{\textbf{value}} by matrix multiplication to get the final results.

The upper right and the lower right in \figref{fig:sparse_dot_product_attention} show examples of normalizing a row vector of the similarity matrix in naive scaled dot-product attention and sparse scaled dot-product attention, respectively.
We can see that naive scaled dot-product attention amplifies relatively smaller similarity weights, which makes the output features susceptible to noises and distractive background.
However, this issue can be significantly alleviated by sparse scaled dot-product attention.

\subsection{Double-Head predictor}\label{sec:double_head_network}
Most existing trackers adopt fully connected network or convolutional network to classification between foreground and background and regression of target bounding boxes, without indepth analysis or design for the head networks based on the characteristics of the tasks of classification and regression.
Inspired by ~\cite{wu2020rethinking}, we introduce a double-head predictor to improve the accuracy of classification and regression.
Specifically, as shown in \figref{fig:architecture}, it consists of a \textit{fc-head} that is composed of two fully connected layers and a \textit{conv-head} that is composed of $L$ convolutional blocks.
Unfocused tasks are added for extra supervision in training.
In the inference phase, for the classification task, we fuse the classification scores outputted by the \textit{fc-head} and the one outputted by the \textit{conv-head}; for the regression task, we only take the predicted offsets outputted by the \textit{conv-head}.

\subsection{Training Loss}
We follow ~\cite{xu2020siamfc++} to generate training labels of classification scores and regression offsets.
In order to train the whole network end-to-end, the objective function is the weighted sum of classification loss and regression loss, as the following:

\begin{equation}\label{eq:objective_function}
\begin{split}
\mathcal{L} &= \omega_{fc}\cdot \left[\lambda_{fc}\mathcal{L}^{class}_{fc} + \left(1 - \lambda_{fc}\right)\mathcal{L}^{box}_{fc}\right] \\
    &+ \omega_{conv}\cdot \left[\left(1 - \lambda_{conv}\right)\mathcal{L}^{class}_{conv} + \lambda_{conv}\mathcal{L}^{box}_{conv}\right]
\end{split}
\end{equation}
where $\omega_{fc}$, $\lambda_{fc}$, $\omega_{conv}$ and $\lambda_{conv}$ are hyper-parameters.
In practice, we set $\omega_{fc}=2.0$, $\lambda_{fc}=0.7$, $\omega_{conv}=2.5$, $\lambda_{conv}=0.8$.
The functions $\mathcal{L}^{class}_{fc}$ and $\mathcal{L}^{class}_{conv}$ are both implemented by focal loss~\cite{lin2017focal}, and the functions $\mathcal{L}^{box}_{fc}$ and $\mathcal{L}^{box}_{conv}$ are both implemented by IoU loss~\cite{yu2016unitbox}.

\section{Experiments}
\subsection{Implementation Details}
\paragraph{Training Dataset.} We use the \verb|train| splits of TrackingNet~\cite{muller2018trackingnet}, LaSOT~\cite{fan2019lasot}, GOT-10k~\cite{huang2019got}, ILSVRC VID~\cite{russakovsky2015imagenet}, ILSVRC DET~\cite{russakovsky2015imagenet} and COCO~\cite{lin2014microsoft} as the training dataset, in addition to the GOT-10k~\cite{huang2019got} benchmark.
We select two frames with a maximum frame index difference of 100 from each video as the target template and the search region.
In order to increase the diversity of training samples, we set the range of random scaling to $\left[\frac{1}{1+\alpha}, 1+\alpha\right]$ and the range of random translation to $\left[-0.2\beta, 0.2\beta\right]$, in which $\alpha=0.3$, $\beta=\sqrt{(1.5w_{t}+0.5h_{t}) \times (1.5h_{t}+0.5w_{t})}$ for the target template, and $\beta=\frac{t \cdot s}{\sqrt{(1.5w_{s}+0.5h_{s}) \times (1.5h_{s}+0.5w_{s})}}$ for the search region.
Here $w_{t}$ and $h_{t}$ are the width and height of the target in the target template, respectively; $w_{s}$ and $h_{s}$ are the width and height of the target in the search region, respectively; $t$ and $s$ are the sizes of the target template and the search region, respectively.
We set $t = 127$ and $s = 289$ in practice.

\paragraph{Model Settings.} We use the tiny version of Swin Transformer~\cite{liu2021swin} (Swin-T) as the backbone $\varphi$.
In the MSA, SMSA, and MCA, the number of heads is set to 8, the number of channels in the hidden layers of FFN is set to 2048, and the dropout rate is set to 0.1.
The number of encoder layers $N$ and the number of decoder layers $M$ are set to 2, and the sparseness $K$ in SMSA is set to 32.
See \secref{subsec:ablation_study} for more discussions about the hyper parameters in the proposed target focus network.
In the \textit{conv-head} of the double-head predictor, the first convolutional block is set to residual block~\cite{he2016deep}, and other $L-1$ ones are set to bottleneck blocks~\cite{he2016deep}, where $L=8$.

\paragraph{Optimization.} We use AdamW optimizer to train our method for 20 epochs.
In each epoch, we sample 600,000 image pairs from all training datasets.
Note that we only sample 300,000 image pairs from the \verb|train| split for the GOT-10k benchmark.
The batch size is set to 32, and the learning rate and the weight decay are both set to $1 \times 10^{-4}$.
After training for 10 epochs and 15 epochs, the learning rate decreases to $1 \times 10^{-5}$ and $1 \times 10^{-6}$, respectively.
The whole training process takes about 60 hours on 4 NVIDIA RTX 2080 Ti GPUs.
Note that the training time of TransT is about 10 days (240 hours), which is $4 \times$ that of our method.

\subsection{Ablation Study}\label{subsec:ablation_study}
\paragraph{The Number of Encoder Layers.} In our method, the encoder is used to enhance the generalization of target template, thus the number of encoder layers is important to our method.
\tabref{tab:ablation_study_num_encoder_layers} lists the performance of our method using different numbers of encoder layers.
Interestingly, the proposed target focus network can still bring comparable performance without the encoder.
As the number increases, the performance gradually improves.
However, when the number of encoder layers is greater than 2, the performance drops.
We argue that excess encoder layers may lead to overfitting of model training.
Therefore, we set the number of encoder layers to 2 in the remaining experiments.
\begin{table}[htbp]
\small
\centering
\vspace{-0.8em}
\begin{tabularx}{0.49\textwidth}{C|CCCC}
\toprule
$N$ & \textbf{0} & \textbf{1} & \textbf{2} & \textbf{3} \\
\midrule
\textbf{AO} & 0.676 & 0.687 & \textbf{0.693} & 0.679 \\
$\textbf{\text{SR}}_{0.5}$ & 0.770 & 0.783 & \textbf{0.791} & 0.770 \\
$\textbf{\text{SR}}_{0.75}$ & 0.627 & 0.634 & \textbf{0.638} & 0.620 \\
\bottomrule
\end{tabularx}
\vspace{-0.8em}
\caption{The performance of our method on the \texttt{test} split of GOT-10k when setting the number of encoder layers to 0, 1, 2, and 3.}
\label{tab:ablation_study_num_encoder_layers}
\vspace{-1.0em}
\end{table}

\paragraph{The Number of Decoder Layers.} We then explore the best setting for the number of decoder layers $M$, as shown in \tabref{tab:ablation_study_num_decoder_layers}.
Similar to $N$, as the number of decoder layers increases, the performance gradually improves when $M$ is not greater than 2.
We also notice that when $M$ equals 3, the performance decreases and the running speed slows down by large margin.
We speculate that it may be caused by overfitting.
Thus, $M$ is set to 2 in the remaining experiments.

\begin{table}[htbp]
\small
\centering
\vspace{-0.8em}
\begin{tabularx}{0.49\textwidth}{C|CCC}
\toprule
$M$ & \textbf{1} & \textbf{2} & \textbf{3} \\
\midrule
\textbf{AO} & 0.672 & \textbf{0.693} & 0.661 \\
$\textbf{\text{SR}}_{0.5}$ & 0.764 & \textbf{0.791} & 0.754 \\
$\textbf{\text{SR}}_{0.75}$ & 0.619 & \textbf{0.638} & 0.610 \\
\midrule
\textbf{FPS} & 40.2 & 39.9 & 37.7 \\
\bottomrule
\end{tabularx}
\vspace{-0.8em}
\caption{The performance of our method on the \texttt{test} split of GOT-10k when setting the number of decoder layers to 1, 2, and 3.}
\label{tab:ablation_study_num_decoder_layers}
\vspace{-1.0em}
\end{table}

\paragraph{The Sparseness $K$ in SMSA.} In SMSA, the sparseness $K$ significantly affects the activation degree of foreground.
Due to the scale variation of targets, a suitable sparseness $K$ ensures good adaptability and generalization at the same time for SMSA.
\tabref{tab:ablation_study_sparseness} shows the impact of different sparseness values on the performance of our method.
Note that when $K=H^{\prime}W^{\prime}$, SMSA becomes naive MSA.
We find that SMSA always brings better performance than MSA in our method, which shows the effectiveness and superiority of SMSA.
When $K$ is 32, Our method achieves the best performance.
Consequently, we set the sparseness $K$ to 32 in our experiments.

\begin{table}[htbp]
\small
\centering
\vspace{-0.8em}
\begin{tabularx}{0.49\textwidth}{C|CCCCCC}
\toprule
$K$ & \textbf{16} & \textbf{32} & \textbf{64} & \textbf{128} & \textbf{256} & $\boldsymbol{H^{\prime}W^{\prime}}$ \\
\midrule
\textbf{AO} & 0.667 & \textbf{0.693} & 0.680 & 0.677 & 0.682 & 0.662 \\
$\textbf{\text{SR}}_{0.5}$ & 0.763 & \textbf{0.791} & 0.777 & 0.771 & 0.780 & 0.754 \\
$\textbf{\text{SR}}_{0.75}$ & 0.611 & \textbf{0.638} & 0.627 & 0.623 & 0.627 & 0.605 \\
\bottomrule
\end{tabularx}
\vspace{-0.8em}
\caption{The performance of our method on the \texttt{test} split of GOT-10k when setting different sparseness values for SMSA, where $H^{\prime}W^{\prime}$ denotes the number of columns of the similarity matrix.}
\label{tab:ablation_study_sparseness}
\vspace{-1.0em}
\end{table}

\subsection{Comparison with the state-of-the-art}
\textbf{LaSOT} is a large-scale long-term dataset with high-quality annotations.
Its \verb|test| split consists of 280 sequences, the average length of which exceeds 2500 frames.
We evaluate our method on the \verb|test| split of LaSOT and compare it with other competitive methods.
As shown in \tabref{tab:lasot_tab}, our method achieves the best performance in terms of success, precision, and normalized precision metrics.
\begin{table}[htbp]
\small
\centering
\vspace{-0.8em}
\begin{tabularx}{0.49\textwidth}{p{0.24\textwidth}<{\centering}CCp{0.07\textwidth}<{\centering}}
\toprule
\textbf{Method} & \textbf{Succ.} & \textbf{Prec.} & \textbf{N. Prec.} \\
\midrule
\textbf{Ours} & \textcolor{red}{\textbf{0.660}} & \textcolor{red}{\textbf{0.701}} & \textcolor{red}{\textbf{0.748}} \\
TransT~\cite{chen2021transformer} & \textcolor{blue}{\textbf{0.649}} & \textcolor{blue}{\textbf{0.690}} & \textcolor{blue}{\textbf{0.738}} \\
TrDiMP~\cite{wang2021transformer} & 0.639 & 0.662 & 0.730 \\
SAOT~\cite{zhou2021saliency} & 0.616 & 0.629 & 0.708 \\
STMTrack~\cite{fu2021stmtrack} & 0.606 & 0.633 & 0.693 \\
DTT~\cite{yu2021high} & 0.601 & - & - \\
AutoMatch~\cite{zhang2021learn} & 0.583 & 0.599 & 0.675 \\
SiamRCR~\cite{PengJGWWTWL21} & 0.575 & 0.599 & - \\
LTMU~\cite{dai2020high} & 0.570 & 0.566 & 0.653 \\
DiMP-50~\cite{bhat2019learning} & 0.565 & 0.563 & 0.646 \\
Ocean~\cite{zhang2020ocean} & 0.560 & 0.566 & 0.651 \\
SiamFC++~\cite{xu2020siamfc++} & 0.543 & 0.547 & 0.623 \\
SiamGAT~\cite{guo2021graph} & 0.539 & 0.530 & 0.633 \\
\bottomrule
\end{tabularx}
\vspace{-0.8em}
\caption{The performance of our method and other excellent ones on the \texttt{test} split of LaSOT, where \enquote{Succ.}, \enquote{Prec.} and \enquote{N. Prec.} represent success, precision and normalized precision, respectively. The best two results are highlighted in \textcolor{red}{\textbf{red}} and \textcolor{blue}{\textbf{blue}}, respectively.}
\label{tab:lasot_tab}
\vspace{-1.0em}
\end{table}

\begin{table*}[t]
\small
\centering
\begin{tabular}{c|p{0.05\textwidth}<{\centering}p{0.05\textwidth}<{\centering}|p{0.05\textwidth}<{\centering}p{0.05\textwidth}<{\centering}|p{0.05\textwidth}<{\centering}p{0.05\textwidth}<{\centering}|p{0.05\textwidth}<{\centering}p{0.05\textwidth}<{\centering}|p{0.05\textwidth}<{\centering}p{0.05\textwidth}<{\centering}}
\toprule
\multirow{2}*{\textbf{Method}} & \multicolumn{2}{c|}{\textbf{Deformation}} & \multicolumn{2}{c|}{\textbf{Partial Occlusion}} & \multicolumn{2}{c|}{\textbf{Scale Variation}} & \multicolumn{2}{c|}{\textbf{Rotation}}  & \multicolumn{2}{c}{\textbf{Viewpoint Change}} \\
\cline{2-11}
~ & \textbf{Succ.} & \textbf{Prec.} & \textbf{Succ.} & \textbf{Prec.} & \textbf{Succ.} & \textbf{Prec.} & \textbf{Succ.} & \textbf{Prec.} & \textbf{Succ.} & \textbf{Prec.} \\
\midrule
\textbf{Ours} & \textcolor{red}{\textbf{0.685}} & \textcolor{red}{\textbf{0.693}} & \textcolor{red}{\textbf{0.634}} & \textcolor{red}{\textbf{0.665}} & \textcolor{red}{\textbf{0.660}} & \textcolor{red}{\textbf{0.700}} & \textcolor{red}{\textbf{0.666}} & \textcolor{red}{\textbf{0.704}} & \textcolor{red}{\textbf{0.673}} & \textcolor{red}{\textbf{0.713}} \\
TransT~\cite{chen2021transformer} & \textcolor{blue}{\textbf{0.670}} & \textcolor{blue}{\textbf{0.674}} & \textcolor{blue}{\textbf{0.620}} & \textcolor{blue}{\textbf{0.650}} & \textcolor{blue}{\textbf{0.646}} & \textcolor{blue}{\textbf{0.687}} & \textcolor{blue}{\textbf{0.643}} & \textcolor{blue}{\textbf{0.687}} & 0.617 & \textcolor{blue}{\textbf{0.654}} \\
TrDiMP~\cite{wang2021transformer} & 0.646 & 0.615 & 0.609 & 0.619 & 0.634 & 0.655 & 0.624 & 0.641 & \textcolor{blue}{\textbf{0.622}} & 0.639 \\
STMTrack~\cite{fu2021stmtrack} & 0.640 & 0.624 & 0.571 & 0.582 & 0.606 & 0.631 & 0.601 & 0.631 & 0.582 & 0.626 \\
SAOT~\cite{zhou2021saliency} & 0.617 & 0.580 & 0.584 & 0.586 & 0.611 & 0.623 & 0.596 & 0.606 & 0.541 & 0.554 \\
AutoMatch~\cite{zhang2021learn} & 0.601 & 0.565 & 0.553 & 0.557 & 0.581 & 0.596 & 0.572 & 0.584 & 0.567 & 0.591 \\
Ocean~\cite{zhang2020ocean} & 0.600 & 0.557 & 0.523 & 0.514 & 0.557 & 0.560 & 0.546 & 0.543 & 0.521 & 0.518 \\
DiMP-50~\cite{bhat2019learning} & 0.574 & 0.506 & 0.537 & 0.516 & 0.560 & 0.554 & 0.549 & 0.533 & 0.553 & 0.568 \\
SiamFC++~\cite{xu2020siamfc++} & 0.574 & 0.532 & 0.509 & 0.497 & 0.544 & 0.546 & 0.548 & 0.549 & 0.514 & 0.538 \\
SiamGAT~\cite{guo2021graph} & 0.571 & 0.509 & 0.512 & 0.485 & 0.540 & 0.530 & 0.538 & 0.527 & 0.500 & 0.498 \\
LTMU~\cite{dai2020high} & 0.560 & 0.494 & 0.530 & 0.511 & 0.565 & 0.558 & 0.543 & 0.528 & 0.587 & 0.599 \\
\bottomrule
\end{tabular}
\vspace{-0.8em}
\caption{The success performance of our method and other excellent ones on the test subsets of LaSOT with attributes of deformation, partial occlusion, scale variation, rotation, and viewpoint change, where \enquote{Succ.} and \enquote{Prec.} represent success and precision, respectively. The best two results are highlighted in \textcolor{red}{\textbf{red}} and \textcolor{blue}{\textbf{blue}}, respectively.}
\label{tab:lasot_attributes_tab}
\end{table*}
We also evaluate our method on the test subsets with attributes of deformation, partial occlusion, and scale variation.
The results are shown in \tabref{tab:lasot_attributes_tab}.
As can be seen, our method performs best in the above challenging scenarios, significantly surpassing other competitive methods.
These challenges bring ambiguous of determining accurate boundaries of targets thus making the trackers hard to locate and estimate target bounding boxes.
However, our method copes with these challenges well.

\textbf{GOT-10k} contains 9335 sequences for training and 180 sequences for testing.
Different from other datasets, GOT-10k only allows trackers to be trained using the \verb|train| split.
We follow this protocol to train our method and test it on the \verb|test| split, then report the performance in \tabref{tab:got10k_tab}.
We see that our method surpasses the second-best tracker TransT by a significant margin, which indicates that our method is superior to other methods when annotated training data is limited.

\begin{table}[htbp]
\small
\centering
\vspace{-0.8em}
\begin{tabularx}{0.49\textwidth}{p{0.27\textwidth}<{\centering}CCC}
\toprule
\textbf{Method} & \textbf{AO} & $\textbf{\text{SR}}_{0.5}$ & $\textbf{\text{SR}}_{0.75}$ \\
\midrule
\textbf{Ours} & \textcolor{red}{\textbf{0.693}} & \textcolor{red}{\textbf{0.791}} & \textcolor{red}{\textbf{0.638}} \\
TransT~\cite{chen2021transformer} & \textcolor{blue}{\textbf{0.671}} & 0.768 & \textcolor{blue}{\textbf{0.609}} \\
  TrDiMP~\cite{wang2021transformer} & 0.671 & \textcolor{blue}{\textbf{0.777}} & 0.583 \\
  AutoMatch~\cite{zhang2021learn} & 0.652 & 0.766 & 0.543 \\
  STMTrack~\cite{fu2021stmtrack} & 0.642 & 0.737 & 0.575 \\
  SAOT~\cite{zhou2021saliency} & 0.640 & 0.749 & - \\
  KYS~\cite{bhat2020know} & 0.636 & 0.751 & 0.515 \\
  DTT~\cite{yu2021high} & 0.634 & 0.749 & 0.514 \\
  PrDiMP~\cite{danelljan2020probabilistic} & 0.634 & 0.738 & 0.543 \\
  SiamGAT~\cite{guo2021graph} & 0.627 & 0.743 & 0.488 \\
  SiamRCR~\cite{PengJGWWTWL21} & 0.624 & - & - \\
  DiMP-50~\cite{bhat2019learning} & 0.611 & 0.717 & 0.492 \\
\bottomrule
\end{tabularx}
\vspace{-0.8em}
\caption{The performance of our method and other excellent ones on the \texttt{test} split of GOT-10k. The best two results are highlighted in \textcolor{red}{\textbf{red}} and \textcolor{blue}{\textbf{blue}}, respectively.}
\label{tab:got10k_tab}
\vspace{-1.6em}
\end{table}

\textbf{UAV123} is a low altitude aerial dataset taken by drones, including 123 sequences, with an average of 915 frames per sequence.
Due to the characteristics of aerial images, many targets in this dataset have low resolution, and are prone to have fast motion and motion blur.
In spite of this, our method is still able to cope with these challenges well.
Thus, as shown in \tabref{tab:uav123_otb2015_tab}, our method surpasses other competitive methods and achieves the state-of-the-art performance on UAV123, which demonstrates the generalization and applicability of our method.

\textbf{OTB2015} is a classical testing dataset in visual tracking.
It contains 100 short-term tracking sequences covering 11 common challenges, such as target deformation, occlusion, scale variation, rotation, illumination variation, background clutters, and so on.
We report the performance of our method on OTB2015.
Although the annotations is not very accurate and it has tended to saturation over recent years, as shown in \tabref{tab:uav123_otb2015_tab}, however, our method still outperforms the excellent tracker TransT~\cite{chen2021transformer} and achieves comparable performance.

\begin{table}[htbp]
\small
\centering
\vspace{-0.8em}
\begin{tabularx}{0.49\textwidth}{p{0.24\textwidth}<{\centering}CC}
\toprule
\textbf{Method} & \textbf{UAV123} & \textbf{OTB2015} \\
\midrule
\textbf{Ours} & \textcolor{red}{\textbf{0.704}} & 0.704 \\
TransT~\cite{chen2021transformer} & \textcolor{blue}{\textbf{0.691}} & 0.694 \\
PrDiMP~\cite{danelljan2020probabilistic} & 0.680 & 0.696 \\
TrDiMP~\cite{wang2021transformer} & 0.675 & \textcolor{blue}{\textbf{0.711}} \\
DiMP-50~\cite{bhat2019learning} & 0.654 & 0.684 \\
STMTrack~\cite{fu2021stmtrack} & 0.647 & \textcolor{red}{\textbf{0.719}} \\
\bottomrule
\end{tabularx}
\vspace{-0.8em}
\caption{The performance of our method and other excellent ones on UAV123 and OTB2015. The best two results are highlighted in \textcolor{red}{\textbf{red}} and \textcolor{blue}{\textbf{blue}}, respectively.}
\label{tab:uav123_otb2015_tab}
\vspace{-1.0em}
\end{table}

\begin{table}[!t]
\small
\centering
\vspace{-0.8em}
\begin{tabularx}{0.49\textwidth}{p{0.24\textwidth}<{\centering}CCp{0.07\textwidth}<{\centering}}
\toprule
\textbf{Method} & \textbf{Succ.} & \textbf{Prec.} & \textbf{N. Prec.} \\
\midrule
\textbf{Ours} & \textcolor{red}{\textbf{81.7}} & \textcolor{blue}{\textbf{79.5}} & \textcolor{blue}{\textbf{86.6}} \\
TransT~\cite{chen2021transformer} & \textcolor{blue}{\textbf{81.4}} & \textcolor{red}{\textbf{80.3}} & \textcolor{red}{\textbf{86.7}} \\
  STMTrack~\cite{fu2021stmtrack} & 80.3 & 76.7 & 85.1 \\
  DTT~\cite{yu2021high} & 79.6 & 78.9 & 85.0 \\
  TrDiMP~\cite{wang2021transformer} & 78.4 & 73.1 & 83.3 \\
  SiamRCR~\cite{PengJGWWTWL21} & 76.4 & 71.6 & 81.8 \\
  AutoMatch~\cite{zhang2021learn} & 76.0 & 72.6 & - \\
  PrDiMP~\cite{danelljan2020probabilistic} & 75.8 & 70.4 & 81.6 \\
  SiamFC++~\cite{xu2020siamfc++} & 75.4 & 70.5 & 80.0 \\
  DiMP-50~\cite{bhat2019learning} & 74.0 & 68.7 & 80.1 \\
\bottomrule
\end{tabularx}
\vspace{-0.8em}
\caption{The performance of our method and other excellent ones on the \texttt{test} split of TrackingNet, where \enquote{Succ.}, \enquote{Prec.} and \enquote{N. Prec.} represent success, precision and normalized precision, respectively. The best two results are highlighted in \textcolor{red}{\textbf{red}} and \textcolor{blue}{\textbf{blue}}, respectively.}
\label{tab:trackingnet_tab}
\vspace{-1.6em}
\end{table}

\textbf{TrackingNet} is a large-scale dataset whose \verb|test| split includes 511 sequences covering various object classes and tracking scenes.
We report the performance of our method on the \verb|test| split of TrackingNet.
As shown in \tabref{tab:trackingnet_tab}, our method achieves the best performance in terms of success metric.

\section{Conclusions}
In this work, we boost Transformer based visual tracking with a novel sparse Transformer tracker.
The sparse self-attention mechanism in Transformer relieves the issue of concentration on the global context and thus negligence of the most relevant information faced by the vanilla self-attention mechanism, thereby highlighting potential targets in the search regions.
In addition, a double-head predictor is introduced to improve the accuracy of classification and regression.
Experiments show that our method can significantly outperform the state-of-the-art approaches on multiple datasets while running at a real-time speed, which demonstrates the superiority and applicability of our method.
Besides, the training time of our method is only 25\% of TransT.
Overall, it is a new excellent baseline for further researches.

\bibliographystyle{named}
\small{\bibliography{ijcai22}}

\begin{thebibliography}{}

\bibitem[\protect\citeauthoryear{Bertinetto \bgroup \em et al.\egroup
  }{2016}]{bertinetto2016fully}
Luca Bertinetto, Jack Valmadre, Joao~F Henriques, Andrea Vedaldi, and Philip~HS
  Torr.
\newblock Fully-convolutional siamese networks for object tracking.
\newblock In {\em ECCV}, pages 850--865, 2016.

\bibitem[\protect\citeauthoryear{Bhat \bgroup \em et al.\egroup
  }{2019}]{bhat2019learning}
Goutam Bhat, Martin Danelljan, Luc~Van Gool, and Radu Timofte.
\newblock Learning discriminative model prediction for tracking.
\newblock In {\em ICCV}, pages 6182--6191, 2019.

\bibitem[\protect\citeauthoryear{Bhat \bgroup \em et al.\egroup
  }{2020}]{bhat2020know}
Goutam Bhat, Martin Danelljan, Luc Van~Gool, and Radu Timofte.
\newblock Know your surroundings: Exploiting scene information for object
  tracking.
\newblock In {\em ECCV}, pages 205--221, 2020.

\bibitem[\protect\citeauthoryear{Carion \bgroup \em et al.\egroup
  }{2020}]{carion2020end}
Nicolas Carion, Francisco Massa, Gabriel Synnaeve, Nicolas Usunier, Alexander
  Kirillov, and Sergey Zagoruyko.
\newblock End-to-end object detection with transformers.
\newblock In {\em ECCV}, pages 213--229, 2020.

\bibitem[\protect\citeauthoryear{Chen \bgroup \em et al.\egroup
  }{2021}]{chen2021transformer}
Xin Chen, Bin Yan, Jiawen Zhu, Dong Wang, Xiaoyun Yang, and Huchuan Lu.
\newblock Transformer tracking.
\newblock In {\em CVPR}, pages 8126--8135, 2021.

\bibitem[\protect\citeauthoryear{Dai \bgroup \em et al.\egroup
  }{2020}]{dai2020high}
Kenan Dai, Yunhua Zhang, Dong Wang, Jianhua Li, Huchuan Lu, and Xiaoyun Yang.
\newblock High-performance long-term tracking with meta-updater.
\newblock In {\em CVPR}, pages 6298--6307, 2020.

\bibitem[\protect\citeauthoryear{Danelljan \bgroup \em et al.\egroup
  }{2020}]{danelljan2020probabilistic}
Martin Danelljan, Luc~Van Gool, and Radu Timofte.
\newblock Probabilistic regression for visual tracking.
\newblock In {\em CVPR}, pages 7183--7192, 2020.

\bibitem[\protect\citeauthoryear{Fan \bgroup \em et al.\egroup
  }{2019}]{fan2019lasot}
Heng Fan, Liting Lin, Fan Yang, Peng Chu, Ge~Deng, Sijia Yu, Hexin Bai, Yong
  Xu, Chunyuan Liao, and Haibin Ling.
\newblock Lasot: A high-quality benchmark for large-scale single object
  tracking.
\newblock In {\em CVPR}, pages 5374--5383, 2019.

\bibitem[\protect\citeauthoryear{Fu \bgroup \em et al.\egroup
  }{2021}]{fu2021stmtrack}
Zhihong Fu, Qingjie Liu, Zehua Fu, and Yunhong Wang.
\newblock Stmtrack: Template-free visual tracking with space-time memory
  networks.
\newblock In {\em CVPR}, pages 13774--13783, 2021.

\bibitem[\protect\citeauthoryear{Guo \bgroup \em et al.\egroup
  }{2021}]{guo2021graph}
Dongyan Guo, Yanyan Shao, Ying Cui, Zhenhua Wang, Liyan Zhang, and Chunhua
  Shen.
\newblock Graph attention tracking.
\newblock In {\em CVPR}, pages 9543--9552, 2021.

\bibitem[\protect\citeauthoryear{He \bgroup \em et al.\egroup
  }{2016}]{he2016deep}
Kaiming He, Xiangyu Zhang, Shaoqing Ren, and Jian Sun.
\newblock Deep residual learning for image recognition.
\newblock In {\em CVPR}, pages 770--778, 2016.

\bibitem[\protect\citeauthoryear{Huang \bgroup \em et al.\egroup
  }{2019}]{huang2019got}
Lianghua Huang, Xin Zhao, and Kaiqi Huang.
\newblock Got-10k: A large high-diversity benchmark for generic object tracking
  in the wild.
\newblock {\em TPAMI}, 2019.

\bibitem[\protect\citeauthoryear{Li \bgroup \em et al.\egroup
  }{2019}]{li2019siamrpn++}
Bo~Li, Wei Wu, Qiang Wang, Fangyi Zhang, Junliang Xing, and Junjie Yan.
\newblock Siamrpn++: Evolution of siamese visual tracking with very deep
  networks.
\newblock In {\em CVPR}, pages 4282--4291, 2019.

\bibitem[\protect\citeauthoryear{Liao \bgroup \em et al.\egroup
  }{2020}]{liaopg}
Bingyan Liao, Chenye Wang, Yayun Wang, Yaonong Wang, and Jun Yin.
\newblock Pg-net: Pixel to global matching network for visual tracking.
\newblock In {\em ECCV}, 2020.

\bibitem[\protect\citeauthoryear{Lin \bgroup \em et al.\egroup
  }{2014}]{lin2014microsoft}
Tsung-Yi Lin, Michael Maire, Serge Belongie, James Hays, Pietro Perona, Deva
  Ramanan, Piotr Doll{\'a}r, and C~Lawrence Zitnick.
\newblock Microsoft coco: Common objects in context.
\newblock In {\em ECCV}, pages 740--755, 2014.

\bibitem[\protect\citeauthoryear{Lin \bgroup \em et al.\egroup
  }{2017}]{lin2017focal}
Tsung-Yi Lin, Priya Goyal, Ross Girshick, Kaiming He, and Piotr Doll{\'a}r.
\newblock Focal loss for dense object detection.
\newblock In {\em ICCV}, pages 2980--2988, 2017.

\bibitem[\protect\citeauthoryear{Liu \bgroup \em et al.\egroup
  }{2021}]{liu2021swin}
Ze~Liu, Yutong Lin, Yue Cao, Han Hu, Yixuan Wei, Zheng Zhang, Stephen Lin, and
  Baining Guo.
\newblock Swin transformer: Hierarchical vision transformer using shifted
  windows.
\newblock In {\em ICCV}, 2021.

\bibitem[\protect\citeauthoryear{Muller \bgroup \em et al.\egroup
  }{2018}]{muller2018trackingnet}
Matthias Muller, Adel Bibi, Silvio Giancola, Salman Alsubaihi, and Bernard
  Ghanem.
\newblock Trackingnet: A large-scale dataset and benchmark for object tracking
  in the wild.
\newblock In {\em ECCV}, pages 300--317, 2018.

\bibitem[\protect\citeauthoryear{Peng \bgroup \em et al.\egroup
  }{2021}]{PengJGWWTWL21}
Jinlong Peng, Zhengkai Jiang, Yueyang Gu, Yang Wu, Yabiao Wang, Ying Tai,
  Chengjie Wang, and Weiyao Lin.
\newblock Siamrcr: Reciprocal classification and regression for visual object
  tracking.
\newblock In {\em IJCAI}, pages 952--958, 2021.

\bibitem[\protect\citeauthoryear{Russakovsky \bgroup \em et al.\egroup
  }{2015}]{russakovsky2015imagenet}
Olga Russakovsky, Jia Deng, Hao Su, Jonathan Krause, Sanjeev Satheesh, Sean Ma,
  Zhiheng Huang, Andrej Karpathy, Aditya Khosla, Michael Bernstein, et~al.
\newblock Imagenet large scale visual recognition challenge.
\newblock {\em IJCV}, 115(3):211--252, 2015.

\bibitem[\protect\citeauthoryear{Vaswani \bgroup \em et al.\egroup
  }{2017}]{Vaswani2017AttentionIA}
Ashish Vaswani, Noam Shazeer, Niki Parmar, Jakob Uszkoreit, Llion Jones,
  Aidan~N Gomez, {\L}ukasz Kaiser, and Illia Polosukhin.
\newblock Attention is all you need.
\newblock In {\em NIPS}, pages 5998--6008, 2017.

\bibitem[\protect\citeauthoryear{Wang \bgroup \em et al.\egroup
  }{2021}]{wang2021transformer}
Ning Wang, Wengang Zhou, Jie Wang, and Houqiang Li.
\newblock Transformer meets tracker: Exploiting temporal context for robust
  visual tracking.
\newblock In {\em CVPR}, pages 1571--1580, 2021.

\bibitem[\protect\citeauthoryear{Wu \bgroup \em et al.\egroup
  }{2020}]{wu2020rethinking}
Yue Wu, Yinpeng Chen, Lu~Yuan, Zicheng Liu, Lijuan Wang, Hongzhi Li, and Yun
  Fu.
\newblock Rethinking classification and localization for object detection.
\newblock In {\em CVPR}, pages 10186--10195, 2020.

\bibitem[\protect\citeauthoryear{Xu \bgroup \em et al.\egroup
  }{2020}]{xu2020siamfc++}
Yinda Xu, Zeyu Wang, Zuoxin Li, Ye~Yuan, and Gang Yu.
\newblock Siamfc++: Towards robust and accurate visual tracking with target
  estimation guidelines.
\newblock In {\em AAAI}, pages 12549--12556, 2020.

\bibitem[\protect\citeauthoryear{Yan \bgroup \em et al.\egroup
  }{2021a}]{Yan_2021_ICCV}
Bin Yan, Houwen Peng, Jianlong Fu, Dong Wang, and Huchuan Lu.
\newblock Learning spatio-temporal transformer for visual tracking.
\newblock In {\em ICCV}, pages 10448--10457, 2021.

\bibitem[\protect\citeauthoryear{Yan \bgroup \em et al.\egroup
  }{2021b}]{yan2021alpha}
Bin Yan, Xinyu Zhang, Dong Wang, Huchuan Lu, and Xiaoyun Yang.
\newblock Alpha-refine: Boosting tracking performance by precise bounding box
  estimation.
\newblock In {\em Proceedings of the IEEE/CVF Conference on Computer Vision and
  Pattern Recognition}, pages 5289--5298, 2021.

\bibitem[\protect\citeauthoryear{Yu \bgroup \em et al.\egroup
  }{2016}]{yu2016unitbox}
Jiahui Yu, Yuning Jiang, Zhangyang Wang, Zhimin Cao, and Thomas Huang.
\newblock Unitbox: An advanced object detection network.
\newblock In {\em ACM MM}, pages 516--520, 2016.

\bibitem[\protect\citeauthoryear{Yu \bgroup \em et al.\egroup
  }{2021}]{yu2021high}
Bin Yu, Ming Tang, Linyu Zheng, Guibo Zhu, Jinqiao Wang, Hao Feng, Xuetao Feng,
  and Hanqing Lu.
\newblock High-performance discriminative tracking with transformers.
\newblock In {\em ICCV}, pages 9856--9865, 2021.

\bibitem[\protect\citeauthoryear{Zhang \bgroup \em et al.\egroup
  }{2020}]{zhang2020ocean}
Zhipeng Zhang, Houwen Peng, Jianlong Fu, Bing Li, and Weiming Hu.
\newblock Ocean: Object-aware anchor-free tracking.
\newblock In {\em ECCV}, 2020.

\bibitem[\protect\citeauthoryear{Zhang \bgroup \em et al.\egroup
  }{2021}]{zhang2021learn}
Zhipeng Zhang, Yihao Liu, Xiao Wang, Bing Li, and Weiming Hu.
\newblock Learn to match: Automatic matching network design for visual
  tracking.
\newblock In {\em ICCV}, pages 13339--13348, 2021.

\bibitem[\protect\citeauthoryear{Zhao \bgroup \em et al.\egroup
  }{2019}]{zhao2019explicit}
Guangxiang Zhao, Junyang Lin, Zhiyuan Zhang, Xuancheng Ren, Qi~Su, and Xu~Sun.
\newblock Explicit sparse transformer: Concentrated attention through explicit
  selection.
\newblock {\em arXiv preprint arXiv:1912.11637}, 2019.

\bibitem[\protect\citeauthoryear{Zhou \bgroup \em et al.\egroup
  }{2021}]{zhou2021saliency}
Zikun Zhou, Wenjie Pei, Xin Li, Hongpeng Wang, Feng Zheng, and Zhenyu He.
\newblock Saliency-associated object tracking.
\newblock In {\em ICCV}, pages 9866--9875, 2021.

\end{thebibliography}

\end{document}